\documentclass[10pt,twocolumn,letterpaper]{article}

\usepackage[pagenumbers]{cvpr} 

\usepackage[dvipsnames]{xcolor}

\definecolor{cvprblue}{rgb}{0.21,0.49,0.74}
\usepackage[pagebackref,breaklinks,colorlinks,citecolor=cvprblue]{hyperref}
\usepackage{multirow}
\usepackage{dsfont}
\usepackage[accsupp]{axessibility}

\makeatletter
\usepackage{tikz}
\newcommand*\circled[2][1.6]{\tikz[baseline=(char.base)]{
    \node[shape=circle, draw, inner sep=1pt, 
        minimum height={\f@size*#1},] (char) {\vphantom{WAH1g}#2};}}
\makeatother

\def\paperID{13360} 
\def\confName{CVPR}
\def\confYear{2024}

\newcommand{\ttb}{\textsc{Task2Box}\xspace}
\newcommand{\ttv}{\textsc{Task2Vec}\xspace}

\title{\ttb: Box Embeddings for Modeling Asymmetric Task Relationships}

\author{
    Rangel Daroya \qquad
    Aaron Sun \qquad
    Subhransu Maji \\
    University of Massachusetts, Amherst \\
    {\tt\small \{rdaroya, aaronsun, smaji\}@umass.edu}
}

\begin{document}

\twocolumn[{%
\renewcommand\twocolumn[1][]{#1}%
\maketitle\begin{center}
    \centering
    \captionsetup{type=figure}
    \vspace{-6mm}
    \includegraphics[scale=0.36]{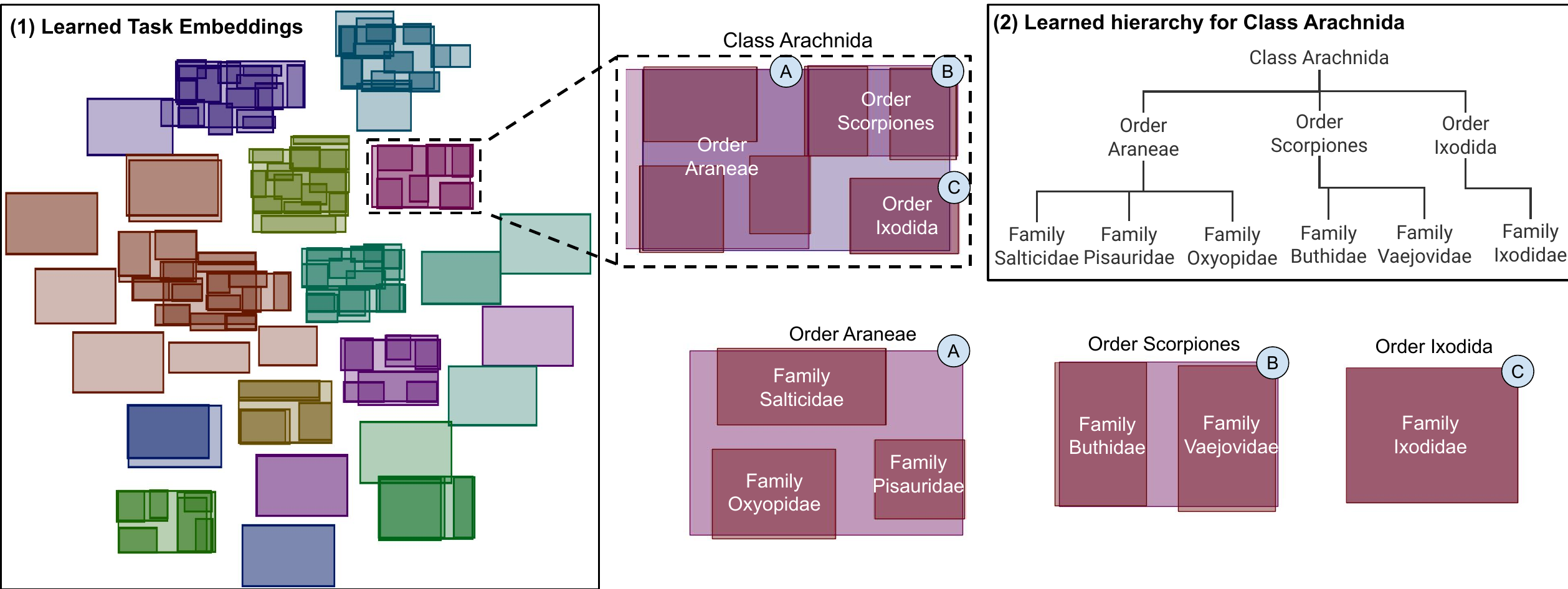}
    \captionof{figure}{\textbf{Box Embeddings of 150 Datasets of iNaturalist + CUB and Corresponding Learned Hierarchy for Class Arachnida}. Each taxonomic category is treated as a separate dataset for which \ttb embeddings are learned. (1) Shows the learned box embeddings where datasets from the same group (taxonomic class) have the same color. Datasets naturally cluster to their ground truth groups. (2) Shows the hierarchy learned through \ttb for a specific class. The hierarchy matches the ground truth relationships based on biological classification. Orders that belong to class Arachnida are learned as boxes (\circled[0]{A}, \circled[0]{B}, \circled[0]{C}) contained by the larger box for Arachnida; families under each of the orders are learned as smaller boxes contained by the corresponding orders they belong to. }
    \label{fig:inatcub-box}
\end{center}%
}]

\begin{abstract}
Modeling and visualizing relationships between tasks or datasets is an important step towards solving various meta-tasks such as dataset discovery, multi-tasking, and transfer learning.
However, many relationships, such as containment and transferability, are naturally asymmetric and current approaches for representation and visualization (e.g., t-SNE \cite{van2008tsne}) do not readily support this.
We propose \ttb, an approach to represent tasks using box embeddings---axis-aligned hyperrectangles in low dimensional spaces---that can capture asymmetric relationships between them through volumetric overlaps.
We show that \ttb accurately predicts unseen hierarchical relationships between nodes in ImageNet and iNaturalist datasets, as well as transferability between tasks in the Taskonomy benchmark. 
We also show that box embeddings estimated from task representations (e.g., CLIP \cite{radford2021clip}, Task2Vec \cite{achille2019task2vec}, or attribute based \cite{gebru2021datasheets}) can be used to predict relationships between unseen tasks more accurately than classifiers trained on the same representations, as well as handcrafted asymmetric distances (e.g., KL divergence). 
This suggests that low-dimensional box embeddings can effectively capture these task relationships and have the added advantage of being interpretable. We use the approach to visualize relationships among publicly available image classification datasets on popular dataset hosting platform called Hugging Face.
\end{abstract}

\section{Introduction}
\label{sec:intro}
The success of deep learning has led to the proliferation of datasets for solving a wide range of computer vision problems. Yet, there are few tools available to enable practitioners to find datasets related to the task at hand, and to solve various meta-tasks related to it. We present \ttb, a method to represent tasks using axis-aligned hyperrectangles (or box embeddings). \ttb is framed as a learnable mapping from dataset representation to boxes, and can be trained to predict various relationships between novel tasks such as transferability, hierarchy, and overlap.

Box embeddings~\cite{vilnis2018probabilistic} extend order embeddings~\cite{vendrov2015order} by using volumetric relationships between axis-aligned hyperrectangles to encode pairwise relationships. Prior work in natural language processing has utilized box embeddings to represent the WordNet \cite{miller1995wordnet} hierarchy and to model conditional distributions. To model relationships between novel datasets, we develop a technique to map from Euclidean representations of datasets into the space of boxes. We explore simple image and label embedding from large vision-language models such as CLIP \cite{radford2021clip}, \ttv \cite{achille2019task2vec}, and attribute-based vectors \cite{gebru2021datasheets} as base representations of tasks.

We test our framework to model asymmetric relationships between nodes in iNaturalist \cite{van2018inaturalist} and Caltech-UCSD Birds (CUB) \cite{cub200_welinder2010} and ImageNet \cite{deng2009imagenet} datasets, as well as to predict transferability on the Taskonomy benchmark \cite{zamir2018taskonomy}.
Table~\ref{table:inat-results} and~\ref{table:imagenet-results} show that low-dimensional box embeddings accurately predict novel relationships between datasets seen during training, as well as relationships with novel datasets. Remarkably, \ttb outperforms classifiers trained to directly predict the relationships on the same representations, suggesting that the box embedding provides a strong inductive bias for learning hierarchical relationships. We also outperform simple asymmetric distances proposed in prior work such as Kullback-Leibler (KL) divergence \cite{achille2019task2vec}. To model the heterogeneous tasks in the Taskonomy benchmark~\cite{zamir2018taskonomy} we map each task to a set of attributes from which a box embedding is learned. Once again, we obtain significantly higher correlation between the true and predicted transferability for both existing and novel datasets compared to standard classifiers (Table \ref{table:affinity-results}). Such attribute-based representations can be readily derived from datasheets \cite{gebru2021datasheets} and modelcards \cite{mitchell2019model}.

Finally, the low-dimensional box embeddings have the added advantage of being interpretable. Fig.~\ref{fig:inatcub-box} and Fig.~\ref{fig:imagenet-box} show relationships on the iNaturalist+CUB and ImageNet categories, respectively. The 2D box representation allows us to readily visualize the strength and direction of task relationships based on the overlapping volumes, which is not possible using symmetric distances with Euclidean representations (e.g., t-SNE \cite{van2008tsne}). At the same time, new datasets can be embedded in constant time without needing to retrain or re-optimize. Fig.~\ref{fig:huggingface} uses \ttb to visualize relationships among 131 publicly available datasets on Hugging Face \cite{huggingfacedatasets}, a popular platform for hosting datasets. Our main contributions are as follows:
\begin{itemize}[noitemsep,topsep=0pt,parsep=0pt,partopsep=0pt,leftmargin=*]
    \item We introduce a novel method (\ttb) that uses box embeddings to learn asymmetric (e.g., hierarchical, transfer learning) dataset relationships.
    \item We demonstrate that \ttb can predict the relationships of \emph{new tasks} with a collection of existing tasks.
    \item We illustrate the interpretability of our model, and the ability to visualize public classification datasets on Hugging Face.
\end{itemize}

\noindent
The code for this project is publicly available at \url{https://github.com/cvl-umass/task2box}.

\section{Related Work}
\label{sec:related-work}
\noindent\textbf{Task Representations.} Given a dataset ${\cal D} = \{(x_i, y_i)\}_{i=1}^n$, consisting of images $x_i \in {\cal X}$ and labels $y_i \in {\cal Y}$, a range of approaches have been proposed for dataset representation. The most straightforward approach involves modeling the distribution of either the images $x$ or the labels $y$ within the dataset independently, using embeddings referred to in prior work as ``domain" and ``label" embeddings~\cite{achille2019task2vec}. To capture the joint dependency between images and labels, \cite{achille2019task2vec} proposed the use of the Fisher Information Matrix (FIM) derived from a ``probe network" trained to minimize a loss function $\ell(\hat{y}, y)$ over the dataset~\cite{liao2018approximate, pennington2018spectrum, karakida2019universal}. This approach leverages the similarity of FIMs to predict task transferability and for model selection. However, the utility of the FIM critically depends on the choice of the probe network and a pre-defined similarity may not accurately represent the various relationships between datasets.

We also investigate the use of vision-language models (VLMs) such as CLIP~\cite{radford2021clip}. This model, trained on a wide range of visual domains, can generalize to tasks involving vision and language data. CLIP features have been effective for image classification~\cite{conde2021clip, Abdelfattah_2023_ICCV}, semantic segmentation~\cite{Liang_2023_CVPR, Lin_2023_CVPR, He_2023_CVPR}, object detection~\cite{Vidit_2023_CVPR, Wu_2023_CVPR, gu2021open}, and even closing domain gaps for performance improvement~\cite{Lai_2023_ICCV, zhu2021mind}. Both images $x \in {\cal X}$ and labels $y \in {\cal Y}$ represented as text, can be mapped into a shared space using the vision encoder ($\phi$) and text encoder ($\psi$) of CLIP, allowing us to model the dataset as a set of image and label embeddings $\{\big(\phi(x_i), \psi(y_i)\big)\}_{i=1}^n$.

We compare FIMs with representations derived from CLIP as base representations for tasks and \emph{learn} box embeddings to model a variety of relations among tasks. \\

\noindent\textbf{Task Relations in Computer Vision.} Understanding the relationships between tasks can lead to efficient solutions to new tasks. Previous work has measured task similarity by using model gradients~\cite{fifty2021efficiently} or based on their learned features \cite{kang2011learning} for grouping tasks for efficient multi-tasking.
Similarly, predicting which pre-trained models will generalize the best on a new dataset could streamline model selection. Taskonomy \cite{zamir2018taskonomy} investigates transfer learning across vision tasks, varying from segmentation to pose estimation, by computing pairwise transfer distances or task affinities. These affinities are calculated by evaluating the extent to which a model trained on a source task generalizes to a target task \cite{dwivedi2019representation, sharma2021instance}, though this process is computationally expensive. \\

\noindent \textbf{Dataset Visualization.} Low-dimensional Euclidean embeddings derived from UMAP \cite{mcinnes2018umap}, t-SNE \cite{van2008tsne}, and LargeVis \cite{tang2016visualizing} are widely used to visualize relationships between datasets. They have been shown to successfully recover clusters of various data modalities by preserving the relationship of each data point with its neighbors~\cite{pmlr-v75-arora18a, achille2019task2vec, sarfraz2022hierarchical}. In low-dimension space, relationships with other data points are defined by their Euclidean distances. However, these are commonly used to represent symmetric relations. 

Visualizations using tidy trees~\cite{reingold1981tidier}, circle packing~\cite{wang2006visualization}, or cone trees~\cite{melancon1998circular} organize asymmetric relations as tree-structured hierarchies in low dimension. However, cyclic data relationships cannot be properly represented for these methods (e.g., when a node has two or more parents).\\

\noindent \textbf{Asymmetric Distances over Datasets.} Kullback-Leibler (KL) divergence between image or label distributions provides a natural way to represent asymmetric distances between datasets. \ttv \cite{achille2019task2vec} computes the similarity between two tasks (e.g., cosine distance), and introduces asymmetry by using the complexity of the first task as a reference. The complexity is measured by the similarity of the task embedding to a ``trivial embedding" (embedding of a task that is easy or has no examples).

Order embeddings on images were first proposed in~\cite{vendrov2015order} to capture tree-structured relationships. Given an dataset of $P={(u,v)}$ drawn from an partially ordered set $(X, \preceq_X)$, they frame the problem as learning a mapping $f:(X, \preceq_X)\rightarrow (Y, \preceq_Y)$ that is order preserving, i.e., $u \preceq_X v \iff f(u) \preceq_Y f(v)$. The reserved product order was used for $\preceq_Y$, i.e., $x \preceq y \iff x_i > y_i, \forall i$. 
Box embeddings~\cite{chheda2021box} generalized this framework by representing points as axis-aligned hyper-rectangles and using volumetric relationships (e.g., intersection over union) to represent asymmetric relations. They used the framework to model conditional distributions and hypernymy relations (e.g., dog ``is a" mammal) on the WordNet graph~\cite{vilnis2018probabilistic, hwang2022event, abboud2020boxe, ren2019query2box, patel2020representing}. 

Hyperbolic spaces provide yet another way to model asymmetric relationships. Examples include the Poincare disk model which uses hyperbolic cosine (cosh) to measure distance between points in a disk. Poincare embeddings have been similarly used to represent WordNet hierarchies \cite{nickel2017poincare} and other relations in general graphs. Hyperbolic representations have also been proposed for representing images for efficient zero-shot learning given a taxonomic structure of the labels \cite{Liu_2020_hyperbolic}.

To the best of our knowledge, no prior work has explored the use of these spaces for representing entire datasets and their effectiveness in capturing various task relationships. We adopt box embeddings in this work due to the effectiveness over alternatives in previous work~\cite{ren2019query2box, abboud2020boxe, patel2022modeling, boratko2021capacity}, ease of visualization, as well as due to open-source libraries for robust learning. However, instead of learning box embeddings directly, we learn mappings from task representations.
\section{\ttb Framework}
\label{sec:method}

We define the problem as follows: given a collection of datasets $\{\mathcal{D}_1, \mathcal{D}_2, \ldots, \mathcal{D}_m \}$, and an asymmetric relationship given the pairwise relationship between datasets $d(\mathcal{D}_i, \mathcal{D}_j) \in [0,1]$, we aim to encode each dataset into a low-dimension space that preserves the relationships between datasets and is interpretable. 

To achieve this, we propose using box embeddings for encoding each of the datasets. This process involves two main steps: (1) deriving the base representation $e$ of each dataset, and (2) learning a model $f_{\theta}: e \rightarrow z$, where $z \in \mathbb{R}^{2\times k}$ represents a $k$-dimensional axis-aligned hyperrectangle (i.e., box), denoted by its lower left and upper right coordinates. These steps are detailed further below.

\subsection{Base Task Representations}
Each dataset $\mathcal{D}=\{(x_i,y_i)\}_{i=1}^n$ is defined as a collection of pairs of images $x_i \in \mathcal{X}$ and labels $y_i \in \mathcal{Y}$. For obtaining a base embedding $e$ for each dataset, we utilize methods such as CLIP \cite{radford2021clip, schuhmann2022laion}, \ttv \cite{achille2019task2vec}, or attribute-based approaches \cite{gebru2021datasheets}. \\

\noindent \textbf{CLIP.} Using a pre-trained CLIP model \cite{radford2021clip, ilharco_gabriel_2021_5143773_openclip, cherti2023reproducibleclip, schuhmann2022laion}, we compute the mean and variance of the individual sample embeddings within each dataset. For each data sample, the image embedding is concatenated with the label embedding, the latter generated from text prompts (e.g., ``A photo of \textsc{[CLS]}"). This concatenation models the joint distribution of images and labels. Eq. \ref{eq:clip_mu} and \ref{eq:clip_var} detail how the mean and variance embeddings are derived, where $[ i, j]$ represents the concatenation of vectors $i$ and $j$, $\phi$ is the vision encoder, and $\psi$ is the text encoder. The covariance is approximated as diagonal for tractability.

\begin{equation}
    \mu_{CLIP} = \frac{1}{N} \sum_{i=1}^N [\phi(x_i), \psi(y_i)]
    \label{eq:clip_mu}
\end{equation}

\begin{equation}
    \sigma_{CLIP}^2 = \frac{1}{N} \sum_{i=1}^N \left( [\phi(x_i), \psi(y_i)] - \mu_{CLIP} \right)^2
    \label{eq:clip_var}
\end{equation}

\noindent The base representation is defined as $e:=\mu_{CLIP} \in \mathbb{R}^{2048}$ or $e:=[\mu_{CLIP}, \sigma^2_{CLIP}] \in \mathbb{R}^{4096}$. A ViT-H/14 \cite{dosovitskiy2020vit} pretrained on LAION-2B \cite{schuhmann2022laion} was used to extract the embeddings.\\

\noindent \textbf{\ttv}~\cite{achille2019task2vec} encodes a dataset using the approximate Fisher Information Matrix (FIM) of a network trained on the given dataset. The FIM represents the importance of parameters in the feature extractor by perturbing the weights $\hat{w}$ of a given probe network with Gaussian noise $\mathcal{N}(0, \Lambda)$. The precision matrix $\Lambda$ is estimated to be close to an isotropic prior $\mathcal{N}(\hat{w}, \lambda^2I)$ while having a good expected error. Eq. \ref{eq:fim} is minimized to find $\Lambda$ where $H$ is the cross entropy loss, $\hat{w}$ are the weights of the network, $\beta$ is the magnitude of the prior, $x \in \mathcal{X}$, and $y \in \mathcal{Y}$.

\begin{multline}
    \mathcal{L}(\hat{w};\Lambda) = \mathbb{E}_{w\sim\mathcal{N}(\hat{w},\Lambda)}[H_{p_w,\hat{p}}p(y|x)] +\\ 
    \beta KL\left( \mathcal{N}(0,\Lambda) \Vert \mathcal{N}(0,\lambda^2I) \right)
    \label{eq:fim}
\end{multline}

The matrix $\Lambda$ provides an estimate of the FIM and is approximated as a diagonal matrix. The diagonal components are used as a base representation of a dataset ($e:=\text{FIM} \in \mathbb{R}^{17024}$). ResNet-34 \cite{heDeepResidualLearning2016} pretrained on ImageNet \cite{deng2009imagenet} is used as the probe network for all datasets.\\

\noindent \textbf{Attribute-based.} A task can be characterized by a set of $t$ binary attributes \cite{gebru2021datasheets} represented as a vector of dimension $t$. Some of the attributes explored for representing tasks are: (1) Is the task generative? (2) Is the task output in 2D? (3) Does the task involve camera pose estimation? Taking these 3 characteristics, for example, we can represent a 2D segmentation task as the vector $e=[0,1,0]$ as a discriminative 2D task that does not need camera poses.

Tasks in Taskonomy \cite{zamir2018taskonomy} involving multiple modalities benefit from this attribute-based representation due to its model independence. This approach also enables generalization to unseen tasks by identifying the presence or absence of various characteristics. Each vision task in Taskonomy is represented with 15 attributes, resulting in vectors $e \in \mathbb{R}^{15}$. The full list of attributes is in Appendix \ref{sec:supp-taskonomy-vector}.

\subsection{Learning Box Embeddings}
From a base task representation $e$, we learn the parameters $\theta$ of a model $f_\theta: e \rightarrow z$ that preserves the asymmetric similarity function $d(\mathcal{D}_i, \mathcal{D}_j)$ between any two datasets $\mathcal{D}_i$ and $\mathcal{D}_j$. In Eq.~\ref{eq:optim}, the learning objective is shown where $\mathcal{L}_E$ is a loss function (mean squared error), $\mathcal{L}_D$ is a distance function, $\mathcal{L}_R$ is a regularization term, and $\lambda$ is a hyperparameter. The embeddings $z_i, z_j \in \mathbb{R}^{2\times k}$ represent the coordinates of the lower left and the upper right corners of the respective $k$ dimemsional boxes, with $f_{\theta}(e_i)=z_i$ and $f_{\theta}(e_j)=z_j$.

\begin{multline}
    \hat{\theta} = \underset{\theta}{\mathrm{argmin}} \sum_{i,j} ( \mathcal{L}_E \left( d(\mathcal{D}_i, \mathcal{D}_j), d_{box}\left(z_i, z_j \right) \right) \\ +\lambda \mathcal{L}_D \left( z_i, z_j \right) + \mathcal{L}_R )
    \label{eq:optim}
\end{multline}

\begin{figure}[!t]
    \begin{center}
    \includegraphics[scale=0.35]{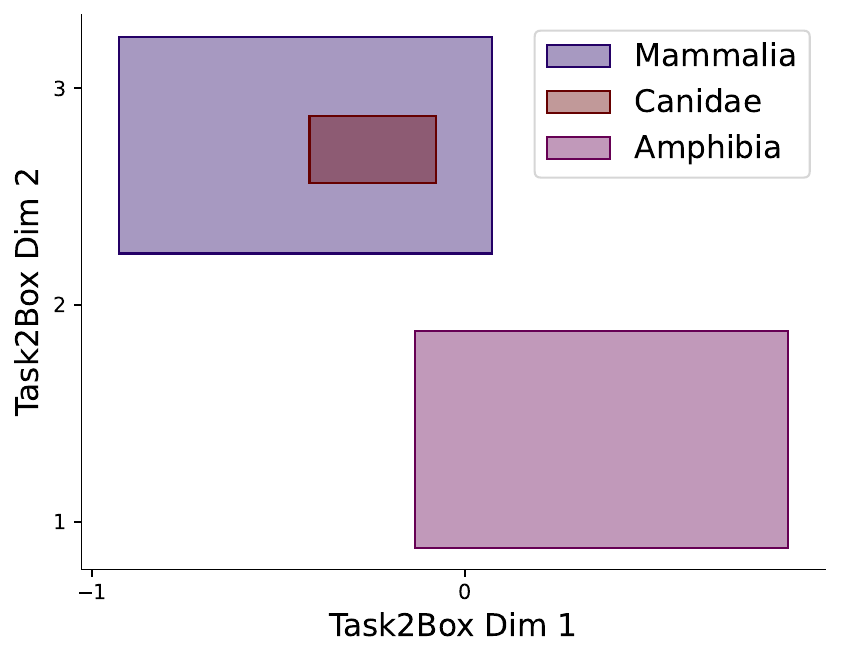}
    \end{center}
    \vspace{-6mm}
    \caption{\textbf{\ttb Embeddings in 2D for Mammalia, Canidae, and Amphibia Datasets from iNaturalist}. Each embedding represents the coordinates of the lower left and upper right corners of each box/rectangle. Since Canidae ($z_1$) is a proper subset of Mammalia ($z_2$): $d_{box}(z_1,z_2)=1$ and $d_{box}(z_2,z_1)=0.1$.}
    \label{fig:mammalia-sample}
    \vspace{-4mm}
\end{figure}

The asymmetric relationship between box embeddings, denoted as $d_{box}(z_i, z_j)$, is computed in Eq.~\ref{eq:box_dist} by calculating the volume of the intersection between $z_i$ and $z_j$, normalized by the volume of $z_i$. For $z_i$ to be fully contained inside $z_j$ ($z_i \subset z_j$), is it required that $d_{box}(z_i, z_j)=1$. Conversely, for $z_j$ to only partially contain $z_i$, $d_{box}(z_j, z_i)$ must fall within the range $(0,1)$. Fig.~\ref{fig:mammalia-sample} illustrates this through a 2-dimensional example, where $z_1$ represents the box embedding for Canidae, and $z_2$ for Mammalia.

\begin{equation}
    d_{box}\left(z_i, z_j \right) = \frac{  \text{vol}(z_i \cap z_j) }{\text{vol}(z_i)}
    \label{eq:box_dist}
\end{equation}

$\mathcal{L}_D$ is applied to datasets where $d(\mathcal{D}_i, \mathcal{D}_j) > 0$ such that the Euclidean distance between the center coordinates of $z_i$ and $z_j$ is minimized when starting from a non-overlapping state. This allows non-overlapping embeddings to move closer to each other, complementing $\mathcal{L}_E$ to learn relationships. $\mathcal{L}_R$ encourages solutions with regular-shaped boxes for better interpretability. The formulation of $\mathcal{L}_R$, given in Eq. \ref{eq:regularizer}, applies to a $k$-dimensional box embedding $z_i$, where $s_a$ represents the size of the $a$-th dimension (for example, width), and $\alpha, \beta$ are hyperparameters. To prevent the trivial solution of minimizing box volume to zero, the inverse of the box volume is included. It's crucial to normalize the terms with respect to the embedding dimension, as the first and second terms scale quadratically and exponentially with dimension increase, respectively.

\begin{equation}
    \mathcal{L}_R = \left( \frac{\alpha}{k^2} \sum_{a=1}^k \sum_{b=a+1}^k |s_a-s_b| \right ) + \beta\left( {\text{vol}(z_i)}\right)^{-1/k}
    \label{eq:regularizer}
\end{equation}

The architecture of $f$ consists of three fully-connected layers followed by two linear heads: one predicts a $k$-dimensional lower-left coordinate, and the other predicts the $k$-dimensional sizes of each box dimension.
\section{Experiments}
\label{sec:experiments}

\begin{table*}[!t]
    \small
    \begin{center}
    \begin{tabular}{l | c c c | c c c}
    \toprule
    \multicolumn{7}{c}{\textbf{iNaturalist + CUB}} \\
    \midrule
    \multirow{2}{*}{\textbf{Method}} & \multicolumn{3}{c|}{\textbf{Existing Datasets}} & \multicolumn{3}{c}{\textbf{Novel Datasets}}\\\cline{2-7}
     & $\mu_{CLIP}$ & $[\mu,\sigma^2]_{CLIP}$ & FIM & $\mu_{CLIP}$ & $[\mu,\sigma^2]_{CLIP}$ & FIM \\
    \midrule
    \ttb (2D) & 69.23\% & 67.84\% & 39.61\% & 50.07\% & 39.66\% & 10.06\% \\
    \ttb (3D) & \underline{79.66\%} & \underline{79.35\%} & \underline{57.63\%} & \underline{70.04\%} & \underline{64.53\%} & \underline{20.65\%} \\
    \ttb (5D) & \textbf{84.67\%} & \textbf{82.41\%} & \textbf{79.72\%} & \textbf{73.79\%} & \textbf{72.11\%} & \textbf{34.88\%} \\
    MLP Classifier & 45.25\% & 61.45\% & 26.34\% & 39.06\% & 44.54\% & 19.90\% \\
    Linear Classifier & 4.40\% & 3.11\% & 7.06\% & 4.77\% & 5.87\% & 15.92\% \\
    \midrule
    KL Divergence & - & 6.58\% & 7.94\% & - & 5.90\% & 0.00\% \\
    Asymmetric Cosine & 9.29\% & 11.54\% & 2.83\% & 1.47\% & 1.47\% & 1.47\% \\
    Asymmetric Euclidean & 1.71\% & 1.71\% & 8.53\% & 1.47\% & 1.47\% & 1.91\% \\
    \midrule
    Random & \multicolumn{3}{c|}{2.06\%} &  \multicolumn{3}{c}{1.49\%}  \\
    \bottomrule
    \end{tabular}
    \caption{\textbf{Average F1 Score for Predicting Hierarchical Relationships on iNaturalist + CUB Dataset.} For all feature types, \ttb outperforms other methods by more than 20\% on existing datasets and more than 10\% on novel datasets. The best-performing model is in \textbf{bold}, and the second-best is \underline{underlined}. Results on \textbf{Novel Datasets} demonstrate that our model can generalize beyond the seen tasks it has previously seen. The dimensions such as 2D, 3D, and 5D refer to different box dimensionalities. Results using KL divergence for $\mu_{CLIP}$ are not shown since it is not a distribution.}
    \label{table:inat-results}
    \end{center}
    \vspace{-6mm}
\end{table*}

We consider the following goals to evaluate the capability of \ttb to represent datasets:
    \begin{enumerate}
        \item Given a collection of \textbf{existing datasets} $\mathcal{D}_E$ and a subset of pairwise relationships $\mathcal{R}$: can the model generalize on unseen relationships $\mathcal{R'}$ within $\mathcal{D}_E$ where $\mathcal{R}' \cap \mathcal{R} = \emptyset$?
        \item Given a collection of \textbf{novel datasets} $\mathcal{D}_N$ not seen during training: can the model accurately identify the relationships with the existing datasets $\mathcal{D}_E$?
    \end{enumerate}
    
\noindent The goals are evaluated through two experimental setups, demonstrating our model's ability to predict various types of relationships. The configurations, and their corresponding datasets, relationships, and metrics, are detailed below. Baseline methods for performance comparison are also reviewed. Implementation details are in the Appendix.


\subsection{Experimental Setup}
\subsubsection{Hierarchical Task Relationships}
\label{subsubsec:hierarchical-def}
We use a combination of iNaturalist \cite{van2018inaturalist} and Caltech-UCSD Birds (CUB) \cite{cub200_welinder2010}, and instrument-related classes in ImageNet \cite{deng2009imagenet} to evaluate the ability of \ttb to represent hierarchical relationships. The first two are composed of images of various species, and the third is composed of images of various objects. The classes naturally follow a hierarchical form based on biological classification (taxonomy of iNaturalist+CUB species), and on semantic relations (hypernymy of objects in ImageNet).

\noindent \textbf{Datasets.}  For iNaturalist+CUB, the datasets are defined as the classes, the orders, and the families in the taxonomy that contain a significant number of samples per dataset as in \cite{achille2019task2vec}. There are 47 classes, 202 orders, and 589 families for a total of 838 datasets. For ImageNet, the instrument-related objects were processed using WordNet \cite{miller1995wordnet} to obtain hierarchical information between classes. This resulted in 131 datasets for training and evaluation. 

\noindent \textbf{Dataset Relationships.} For any two datasets, their relationship is captured as $d(\mathcal{D}_i, \mathcal{D}_j) \in \{0,1\}$, where $d(\mathcal{D}_i, \mathcal{D}_j)=1$ if and only if $\mathcal{D}_i \subset \mathcal{D}_j$, and $d(\mathcal{D}_i, \mathcal{D}_j)=0$ otherwise. Fig.~\ref{fig:mammalia-sample} provides an example: Canidae ($\mathcal{D}_1$) is a family within the class Mammalia ($\mathcal{D}_2$); thus, $d(\mathcal{D}_1, \mathcal{D}_2)=1$ and $d(\mathcal{D}_2, \mathcal{D}_1)=0$. Meanwhile, Amphibia ($\mathcal{D}_3$) is unrelated to either dataset, resulting in $d(\mathcal{D}_1, \mathcal{D}_3)=0$
 

\noindent \textbf{Evaluation Metrics.} We evaluate the ability of the model to classify the presence of containment relationships between datasets using the F1 score due to the imbalance between positive and negative relationships. To obtain F1 score, precision and recall are first calculated. Precision is computed as the ratio of true positive pairs predicted to the total number of positive predictions. Recall is the ratio of true positive pairs predicted to the total number of true positive pairs. F1 score is then reported as the harmonic mean between the precision and recall. These calculations are done on both (1) the unseen relationships $\mathcal{R}'$ within existing datasets $\mathcal{D}_E$ and (2) the relationships between novel datasets $\mathcal{D}_N$ and existing datasets $\mathcal{D}_E$. For the latter, we evaluate the relationships in both directions, i.e., $\{(\mathcal{D}_e, \mathcal{D}_n) \  \forall \ \mathcal{D}_e \in \mathcal{D}_E \} \cup \{(\mathcal{D}_n, \mathcal{D}_e) \  \forall \ \mathcal{D}_e \in \mathcal{D}_E \}$.

\begin{table*}[!t]
    \small
    \begin{center}
    \begin{tabular}{l | c c c | c c c}
    \toprule
    \multicolumn{7}{c}{\textbf{ImageNet}} \\
    \midrule
    \multirow{2}{*}{\textbf{Method}} & \multicolumn{3}{c|}{\textbf{Existing Datasets}} & \multicolumn{3}{c}{\textbf{Novel Datasets}}\\\cline{2-7}
     & $\mu_{CLIP}$ & $[\mu,\sigma^2]_{CLIP}$ & FIM & $\mu_{CLIP}$ & $[\mu,\sigma^2]_{CLIP}$ & FIM \\
    \midrule
    \ttb (2D) & 62.72\% & 63.33\% & 31.32\% & 47.76\% & 48.35\% & 9.46\% \\
    \ttb (3D) & \underline{83.12\%} & \underline{82.79\%} & \underline{58.16\%} & \underline{73.06\%} & \underline{66.48\%} & 24.70\% \\
    \ttb (5D) & \textbf{90.58\%} & \textbf{88.48\%} & \textbf{64.91\%} & \textbf{76.95\%} & \textbf{78.84\%} & \underline{37.39\%} \\
    MLP Classifier & 54.20\% & 60.43\% & 44.85\% & 62.24\% & 64.56\% & \textbf{41.22\%} \\
    Linear Classifier & 9.84\% & 8.33\% & 25.46\% & 11.89\% & 11.87\% & 22.98\% \\
    \midrule
    KL & - & 5.53\% & 9.90\% & - & 10.41\% & 0.00\% \\
    Asymmetric Cosine & 4.28\% & 4.28\% & 6.92\% & 4.52\% & 4.52\% & 0.00\% \\
    Asymmetric Euclidean & 3.73\% & 3.73\% & 7.16\% & 4.52\% & 4.52\% & 4.73\% \\
    \midrule
    Random & \multicolumn{3}{c|}{3.64\%} &  \multicolumn{3}{c}{5.02\%}  \\
    \bottomrule
    \end{tabular}
    \caption{\textbf{Average F1 Score for Predicting Hierarchical Relationships on ImageNet Instruments}. \ttb is shown to generalize well on both predicting relationships between existing datasets (\textbf{Existing Datasets}), and predicting box embeddings of new datasets and their relationships with existing datasets (\textbf{Novel Datasets}).}
    \label{table:imagenet-results}
    \end{center}
    \vspace{-6mm}
\end{table*}
\subsubsection{Transfer Learning Between Datasets}

The Taskonomy~\cite{zamir2018taskonomy} benchmark is used to evaluate the ability of \ttb to predict task affinities. 

\noindent \textbf{Datasets.} Taskonomy defines a set of 25 visual tasks with corresponding pairwise task affinities. These tasks range from object detection, semantic segmentation, pose estimation, and more. We treat each visual task as a dataset. The tasks are described in Appendix~\ref{sec:taskonomy-task-desc}.

\noindent \textbf{Dataset Relationships.} In contrast to the containment relationship defined in \S~\ref{subsubsec:hierarchical-def}, Taskonomy quantifies relationships with task affinity measured by the performance gain achieved by transfer learning from a source dataset $\mathcal{D}_j$ to a target dataset $\mathcal{D}_i$. The values are computed using Analytic Hierarchy Process~\cite{saaty1987analytic, zamir2018taskonomy}. Dataset relationships are computed and normalized based on an ordinal approach and determined by the percentage of images that transfer well to a target task given a set of source tasks as detailed in \cite{zamir2018taskonomy}. For a pair of datasets $(\mathcal{D}_i, \mathcal{D}_j)$, the task affinity is defined as $d(\mathcal{D}_i, \mathcal{D}_j) \in [0,1]$. The higher the task affinity to a target dataset $\mathcal{D}_i$ from a source dataset $\mathcal{D}_j$, then $d(\mathcal{D}_i, \mathcal{D}_j)$ gets closer to 1, where a value of 1 would show $\mathcal{D}_i \subset \mathcal{D}_j$.


\noindent \textbf{Evaluation Metrics.} We evaluate the prediction of the model based on the Spearman correlation between the ground truth task affinity values, and the predicted values based on the box distances in Eq.~\ref{eq:box_dist}. Similar to \S~\ref{subsubsec:hierarchical-def}, we evaluate on both (1) unseen relationships $\mathcal{R}'$ for existing datasets, and (2) on relationships between unseen novel datasets $\mathcal{D}_N$ and existing datasets $\mathcal{D}_E$.

\subsection{Baseline Methods}
 We compare the performance of \ttb with alternative models and simple asymmetric distances proposed in prior work. For hierarchical relationships, given two datasets $\mathcal{D}_i, \mathcal{D}_j$, the models predict $\hat{d}(\mathcal{D}_i, \mathcal{D}_j) \in \{0,1\}$. For task affinity, the models predict $\hat{d}(\mathcal{D}_i, \mathcal{D}_j) \in [0,1]$. The structure for the various methods are discussed below.
 
\noindent\textbf{Linear Model.} A linear model is trained to predict the relationship value between two datasets. The input is the concatenation of the base representation $e$ for the two datasets.

\noindent\textbf{MLP Model.} A 4-layer MLP is used instead for prediction on the same inputs as the linear model.

\noindent \textbf{KL Divergence.} Each dataset is treated as a multivariate Gaussian using the mean and variance of the image and label features (CLIP) or directly as the FIM. The optimal threshold $t$ is selected for the minimum distance between two dataset distributions $\text{KL}(\mathcal{D}_i || \mathcal{D}_j) < t$ for a prediction of $\hat{d}(\mathcal{D}_i, \mathcal{D}_j)=1$, and $\hat{d}(\mathcal{D}_i, \mathcal{D}_j)=0$ otherwise. The F1 Score (hierarchical) or the correlation (task affinity) is the objective for selecting $t$ on the train set. 

\noindent \textbf{Asymmetric Cosine Similarity.} The cosine similarity $d_{cos}$ is a symmetric measure between two embeddings $e_i, e_j$. An asymmetric variant was proposed in \cite{achille2019task2vec} and shown in Eq.~\ref{eq:asym-cos} by considering the similarity of $e_i$ and $e_j$ relative to the complexity of $e_i$. The complexity of is measured as the distance to the trivial embedding $e_o$, and $\alpha$ is a hyperparameter.
\begin{equation}
    d_{asym}(e_i, e_j) = d_{cos}(e_i, e_j) - \alpha d_{cos}(e_i, e_o)
    \label{eq:asym-cos}
\end{equation}

\noindent An optimal threshold $t$ is found on the train set where $d_{asym}(e_i, e_j) < t $ results to a prediction $\hat{d}(\mathcal{D}_i, \mathcal{D}_j)=1$. The same threshold $t$ is used for test set evaluation.

\noindent \textbf{Asymmetric Euclidean Similarity.} The asymmetric similarity is computed as in Eq.~\ref{eq:asym-cos} but uses the Euclidean distance instead of cosine.

\noindent\textbf{Random.} The probability of containment (for hierarchical) or the value of the task affinity is uniformly random.

\section{Results and Discussion}

\label{sec:results}
\subsection{Hierarchical Relationships}

\begin{figure}[!t]
    \begin{center}
    \includegraphics[scale=0.42]{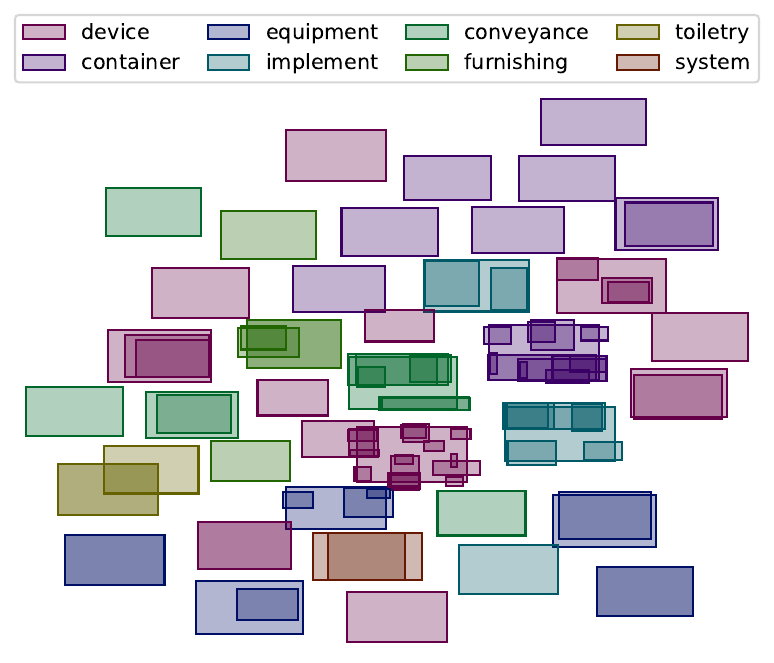}
    \end{center}
    \vspace{-6mm}
    \caption{\textbf{Visualization of Instrument-related Datasets in ImageNet}. Datasets that belong to the same superset are shaded in the same color. \ttb learns the hierarchy of various groups, and clusters similar datasets closer.
    }
    \label{fig:imagenet-box}
    \vspace{-4mm}
\end{figure}
Tables~\ref{table:inat-results} and \ref{table:imagenet-results} present the average F1 score for predicting hierarchical relationships on iNaturalist+CUB and ImageNet datasets. Various methods of extracting dataset embeddings were evaluated. \ttb outperforms baseline methods, indicating that \ttb can both generalize across unseen relationships (Existing Datasets), and to accurately represent relationships with existing datasets for unseen datasets (Novel Datasets). CLIP features also generalize better than FIM features, perhaps because the CLIP multi-modal embedding of images and labels was trained across a broad set of domains\cite{radford2021clip}. However, while CLIP is confined to image and text modalities, FIM can accommodate any modality by adapting the probe network’s last layer for different prediction tasks. 

Beyond predicting relationships, our method allows the visualization of datasets. Fig.~\ref{fig:inatcub-box} illustrates this for a subset of datasets in iNaturalist and CUB using \ttb (2D), and Fig.~\ref{fig:imagenet-box} for ImageNet. The hierarchical organization of the datasets is apparent in the \ttb representation -- datasets that contain others appear as larger boxes, while more specialized datasets are depicted as smaller boxes neted with borader, more general dataset boxes. However, while 2D visualization offers insights, the flat surface may restrict the representation of complex relationships. Representations in higher dimensions is explored in \S~\ref{subsec:analysis}.

\subsection{Task Affinity}
Table~\ref{table:affinity-results} presents the results of task representations in Taskonomy. Our method show that, even with attribute-based embeddings, it can learn box embeddings through relationship supervision between datasets. \ttb is shown to correlate highly with the ground truth task affinities compared to other methods. Given that only 25 tasks are available in Taskonomy, with 3 held out of training as novel datasets, there is a higher uncertainty in predicting unseen datasets. We expect that performance on novel datasets will improve with more datasets available during training.

\begin{figure}[!t]
    \begin{center}
    \includegraphics[scale=0.38]{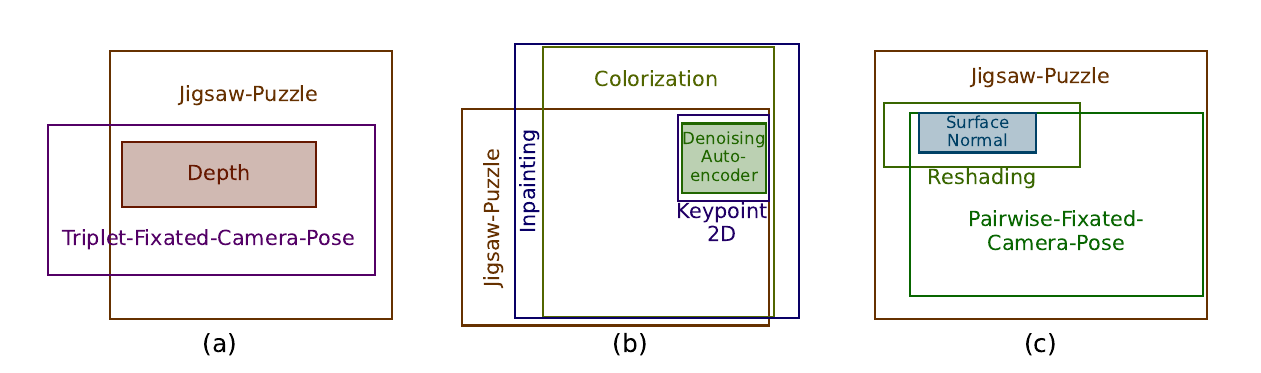}
    \end{center}
    \vspace{-6mm}
    \caption{\textbf{Visualization of Tasks in Taskonomy showing Source Tasks that Transfer Well to Target Tasks (shaded)}. (a) Jigsaw and Triplet Fixated Camera Pose estimation are source tasks that transfer well to Depth estimation. (b) and (c) show different source tasks (larger boxes) that transfer well to the shaded boxes of Denoising Autoencoder and Surface Normal, respectively.
    }
    \label{fig:taskonomy-box}
    \vspace{-3mm}
\end{figure}

Fig.~\ref{fig:taskonomy-box} displays the learned representation with \ttb (2D). Each subfigure in (a)-(c) represents a subset of tasks identified as having strong transfer relationships. \ttb not only identifies but also visually represents related tasks suitable for fine-tuning. The small highlighted boxes indicate target tasks, while the larger enclosing boxes represent source tasks. Although other tasks may not be proper subsets, the box distance in Eq.~\ref{eq:box_dist} provides an estimate for task affinity. A small box distance between two datasets, $d_{box}(z_1, z_2)$, suggests lower transfer performance from a source task $z_2$ to a target task $z_1$.

\begin{table}[!t]
    \small
    \begin{center}
    \begin{tabular}{l | c | c}
    \toprule
    \multirow{2}{*}{\textbf{Method}} & {\textbf{Existing Datasets}} & {\textbf{Novel Datasets}}\\\cline{2-3}
     & Spearman's $\rho$ & Spearman's $\rho$ \\
    \midrule
    \ttb (2D) & 0.85 $\pm$ 0.06 & 0.12 $\pm$ 0.21 \\
    \ttb (3D) & \underline{0.93 $\pm$ 0.02} & \textbf{0.48 $\pm$ 0.24} \\
    \ttb (5D) & \textbf{0.94 $\pm$ 0.03} & 0.39 $\pm$ 0.22\\
    MLP & 0.88 $\pm$ 0.06 & 0.31 $\pm$ 0.18 \\
    Linear & 0.75 $\pm$ 0.11 & \underline{0.40 $\pm$ 0.24} \\
    \midrule
    Random & 0.05 $\pm$ 0.14 &  0.15 $\pm$ 0.07 \\
    \bottomrule
    \end{tabular}
    \caption{\textbf{Spearman Correlation and Standard Deviation between Predicted and Ground Truth Task Affinities on Taskonomy}. Our method shows higher correlation with the task affinities compared to the baseline. Attribute-based embeddings were used.}
    \label{table:affinity-results}
    \end{center}
    \vspace{-6mm}
\end{table}

\begin{figure*}[!t]
    \begin{center}
    \includegraphics[scale=0.38]{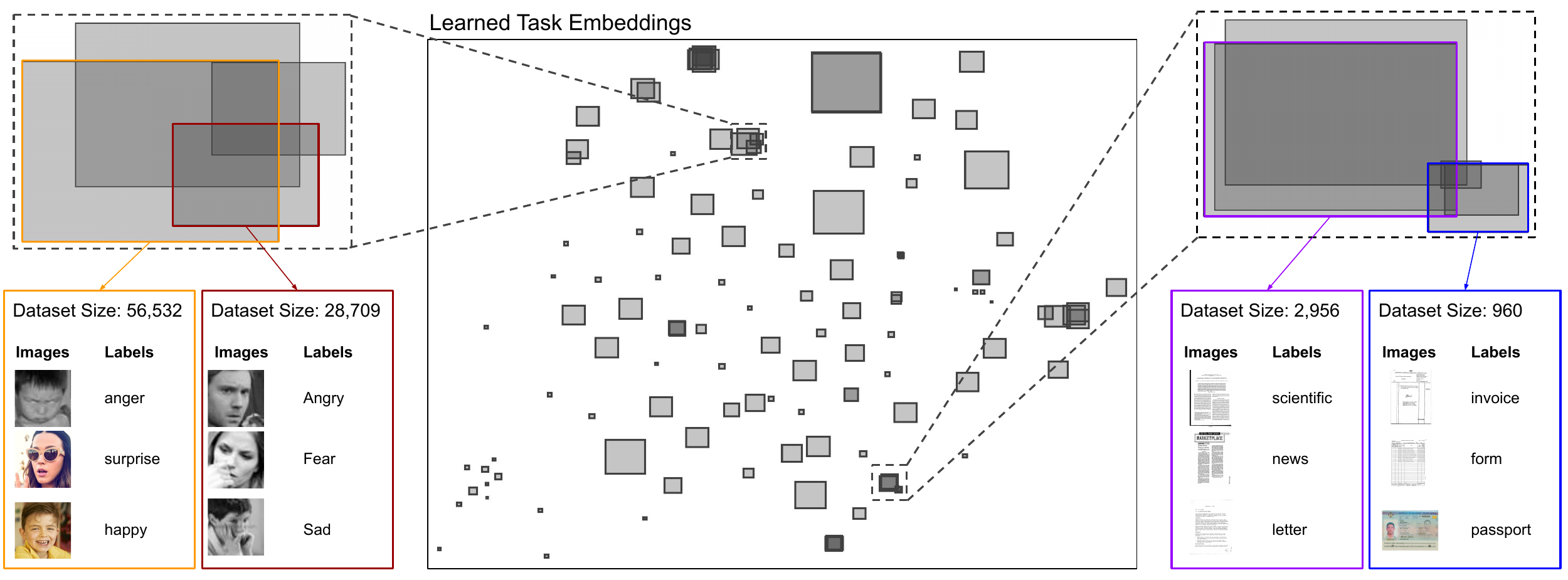}
    \end{center}
    \vspace{-6mm}
    \caption{\textbf{Visualizing Image Classification Datasets in Hugging Face}. The sample data points annotated on the highlighted datasets show that common tasks overlap with each other (e.g., sentiment classification and document classification datasets). Although labels could slightly differ between datasets, \ttb can infer the level of similarity and represent it as the amount of overlap. The embedding size (box area) also shows the number of available data.
    }
    \label{fig:huggingface}
    \vspace{-3mm}
\end{figure*}

\begin{figure}[!t]
    \begin{center}
    \includegraphics[scale=0.38]{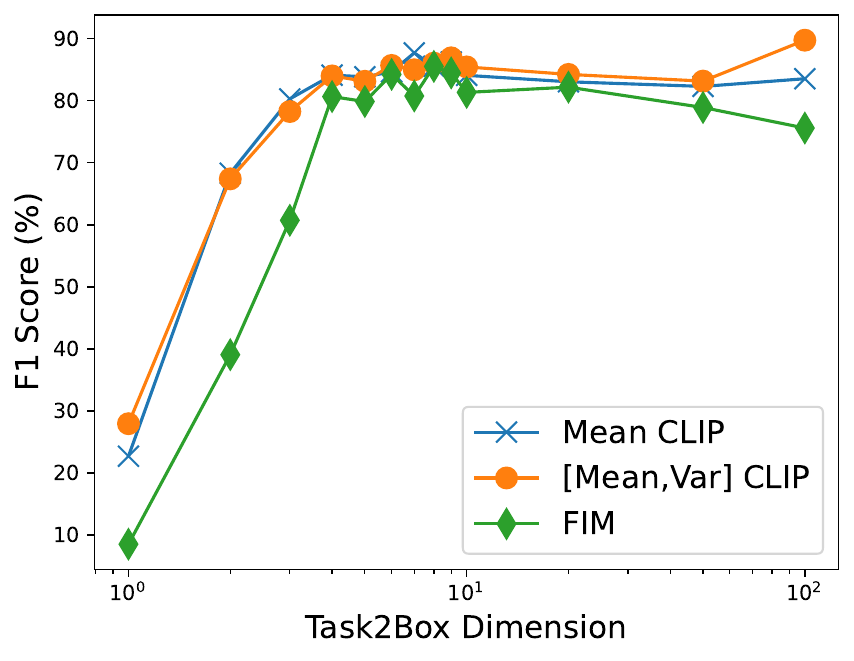}
    \end{center}
    \vspace{-6mm}
    \caption{\textbf{Effect of \ttb Embedding Dimension on the Accuracy of Predicting Task Relations.} As the dimension increases, the performance generally increases.
    }
    \vspace{-4mm}
    \label{fig:boxdim}
    
\end{figure}

\subsection{Visualizing Public Datasets}
Apart from predicting both hierarchical relationships and quantified task affinities, our method can also be used to visualize public datasets that lack available ground truth relationships. Using a set of vision tasks from Hugging Face \cite{huggingfacedatasets}, $\mu_{CLIP}$ + \ttb (2D) were utilized to predict dataset embeddings. A constraint on the box sizes was added to reflect information about the dataset sizes. Fig.~\ref{fig:huggingface} displays the results of \ttb on 131 datasets from Hugging Face, where similar datasets, such as sentiments and documents, are shown to overlap. This approach allows for the analysis and visualization of dataset variations and similarities even with only samples of images and labels. Simultaneously, the sizes of various datasets can be visualized as box sizes -- with larger datasets containing more samples depicted as larger boxes. This tool enables computer vision practitioners to see how their datasets compare with other existing datasets. Beyond visualizing the variation of available datasets (based on the number of clusters), it can expedite the process of finding suitable data sources by examining embedding overlaps.


\subsection{Analysis of \ttb}
\label{subsec:analysis}
We discuss properties of \ttb below. Additional insights are also included in Appendix \ref{sec:supp-addtl-analysis}.

\noindent \textbf{Using a Box Prior Improves Performance.} While \ttb was trained with a 3-layer MLP, it achieved significantly better performance than the MLP baseline without a box prior. Tables \ref{table:inat-results}, \ref{table:imagenet-results}, and \ref{table:affinity-results} demonstrate our method outperforming the baselines. Representing each dataset as an entity with shape and volume (such as a box), instead of solely learning relationships from embeddings, proves effective for generalization on unseen relationships. This improvement could be attributed to the explicit modeling of relationships through physical shapes, which enforces consistency: if $z_i \subset z_j$ and $z_j$ has no overlap with $z_k$, then having a physical representation ensures that $z_i \cap z_k = \emptyset$.

\noindent \textbf{\ttb can Represent Relations in Varying Dimensions.} Fig.~\ref{fig:boxdim} illustrates the performance of \ttb as the box dimension increases. Representing relationships in two dimensions results in easily interpretable embeddings; however, modeling complex task relationships could benefit from expansion to higher dimensions. Projecting to higher dimensions may yield even better performance, as relationships between datasets become more accurately represented, given the model's increased representation capacity. Nonetheless, as the dimensionality further increases, learning and visualizing the embedding space also becomes more challenging.

\noindent \textbf{Using Boxes to Represent Tasks Enables Effective Calculations on Embeddings.} Boxes offer the advantage of being closed under intersection, meaning the intersection of two boxes results in another box, a property not shared by circles or ellipses. Among region-based embeddings that utilize geometric objects (e.g., cones, disks, and boxes) to represent entities, operations involving boxes are the most straightforward to calculate \cite{dasgupta2022word2box, ren2019query2box}.

\section{Conclusion}
\label{sec:conclusion}
We present a novel method that learns low-dimensional, interpretable embeddings of various task relationships. With box representations, we demonstrate that asymmetric relationships, such as task affinities and hierarchies, can be accurately modeled using various base representations. While CLIP embeddings have been shown to outperform FIM when integrated with our model, future work could investigate how CLIP might be adapted to modalities beyond text and images. Attribute-based features offer a viable alternative, and can be extracted from from datasheets using natural language processing techniques.

The distinct properties of \ttb were analyzed, revealing its ability to perform effectively across varying dimensions, and to model and visualize overlaps among public classification datasets on Hugging Face. This could enable computer vision practitioners to assess dataset utility for a task in hand. Although our model successfully represents task relationships, it does not incorporate information about optimal training procedures and model architecture. Future work could explore the inclusion of additional information for a more detailed understanding of the task space.

\paragraph{Acknowledgements.} This work was supported by awards from the National Science Foundation (2329927 and 1749833) and the NASA AIST program. The experiments were performed on the University of Massachusetts GPU cluster funded by the Mass. Technology Collaborative.

{
    \small
    \bibliographystyle{ieeenat_fullname}
    \bibliography{main}
}

\clearpage
\setcounter{page}{1}
\maketitlesupplementary

\section{Further Analysis of \ttb}
\label{sec:supp-addtl-analysis}

\begin{figure}[!t]
    \begin{center}
    \includegraphics[scale=0.37]{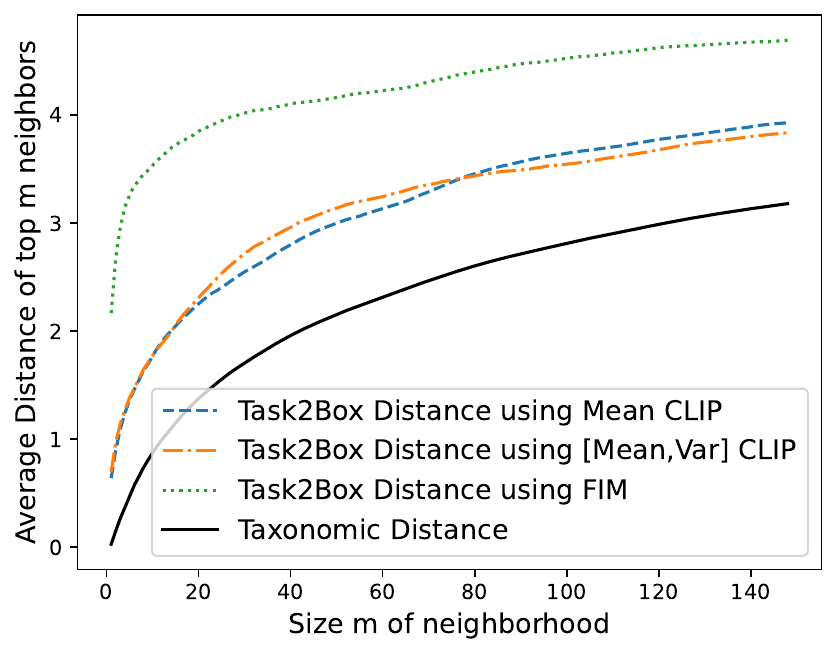}
    \end{center}
    \vspace{-6mm}
    \caption{\textbf{Euclidean Distance between Predicted Embeddings Reflect Taxonomic Distance in iNaturalist+CUB.} \ttb embeddings are projected closer for smaller taxonomic distances. 
    }
    \vspace{-3mm}
    \label{fig:taxdist}
\end{figure}

\noindent \textbf{Euclidean Distance Between \ttb Embeddings Reflects Taxonomic Structure}. Fig.~\ref{fig:taxdist} displays the average embedding and taxonomic distance within a neighborhood of the $m$ closest datasets. Taxonomic distance is defined as the symmetric graph distance in the taxonomy tree. As the number of neighbors $m$ being considered increases, we plot the average distance between \ttb embeddings. Although the model is not directly supervised to mirror taxonomic distances, it inherently learns to position similar datasets closer together.

\begin{figure}[t]
    \begin{center}
    \includegraphics[scale=0.4]{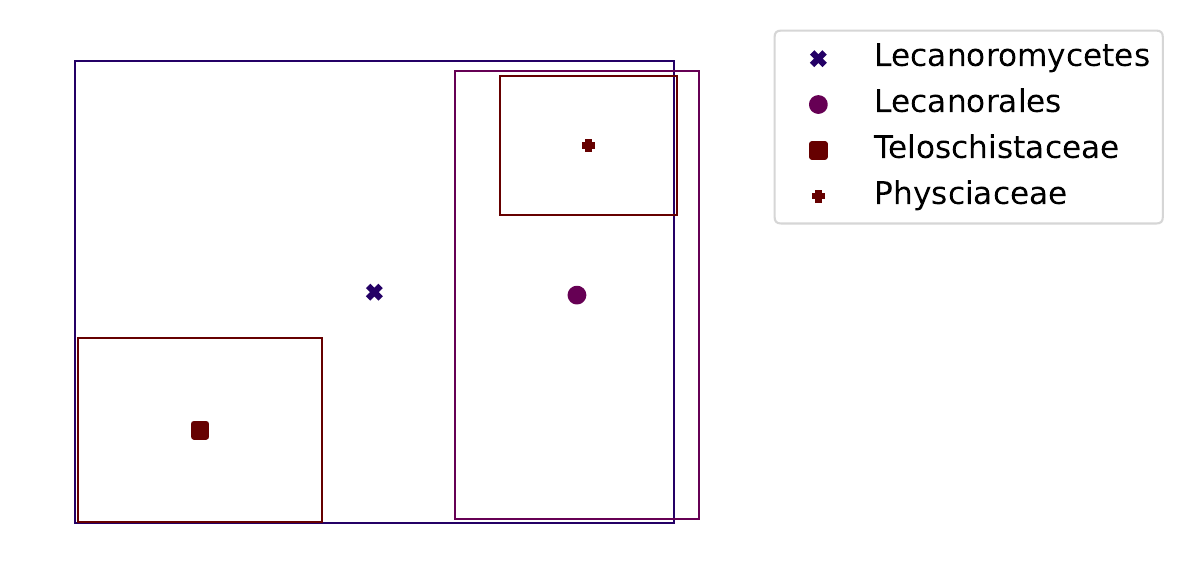}
    \end{center}
    \vspace{-0.4cm}
    \caption{\textbf{Average embeddings (center box coordinates) of superset datasets  are outside the bounds of the subset datasets}. Due to the distinct location of the mean, simply providing the average embedding to the model can be sufficient for training.
    }
    \label{fig:clip-ave}
\end{figure}

\noindent \textbf{Average Dataset CLIP Embeddings are Effective.} While employing both the mean and variance could assist in estimating the general location and spread, Table~\ref{table:inat-results} indicates that using solely the mean across different data points in the dataset generally yields better or equally effective performance. This can be attributed to the unique positioning of the average embedding depending on the hierarchical level (i.e., class, order, or family in iNaturalist). Fig.~\ref{fig:clip-ave} displays box embeddings at various hierarchical levels in iNaturalist, where the center coordinates of supersets lie beyond the bounds of the box representations of their subsets.

\noindent \textbf{\ttb Performs Well Even When Trained on Only 50\% of Relationships Between Datasets.} Table~\ref{table:50perc-supervision} demonstrates the model's ability to generalize to unseen relationships when trained with only 50\% of the pairwise relationships between datasets. Despite the limited relationship data used for training, \ttb can effectively generalize to other relationships. This aspect is particularly valuable in scenarios where acquiring comprehensive dataset relationship information is resource-intensive. For instance, in computing task affinities as done in Taskonomy~\cite{zamir2018taskonomy}, determining transfer learning gains from a source task to a target task necessitates model training. The ability to capture unseen relationships without complete information makes resource-intensive analyses more practical.

\noindent \textbf{\ttb has Several Advantages Over Simple Dimensionality Reduction.} \ttb can predict and visualize asymmetric relationships of \emph{new tasks} in relation to a collection of other existing tasks. Instead of re-optimizing the representation each time a new task is introduced, \ttb predicts an embedding that encapsulates the relationship of the new entity with existing entities. This ability to predict relationships between novel tasks demonstrates that our model can \emph{generalize} beyond the tasks it has previously seen. While existing dimensionality reduction methods such as t-SNE~\cite{van2008tsne} and UMAP~\cite{mcinnes2018umap} provide visualizations, they inherently represent each entity as a point in Euclidean space, resulting in \emph{symmetric} visualizations. In contrast, \ttb can represent \emph{asymmetric} relationships, such as hierarchies and transferability. 

\noindent \textbf{\ttb Can Be Used for Several Applications Related to Task Relationships.} \ttb serves as a tool for visualizing relationships among large collections of datasets, facilitating dataset discovery, measuring overlaps, predicting transferability, and organizing tasks for multitasking. For instance, upon encountering a new task, \ttb can predict its relationship with existing tasks (see Tables~\ref{table:inat-results}, \ref{table:imagenet-results}, and \ref{table:affinity-results} under Novel Datasets). By utilizing task affinities, solutions for the new task can be developed by leveraging models from existing tasks with strong relationships, thereby enabling effective transfer learning. Furthermore, hierarchical relationships can identify which datasets have sufficient overlap to be utilized for further training or evaluation purposes. At present, there are limited techniques available that effectively visualize asymmetric relationships between datasets.

\noindent \textbf{Training \ttb is Most Effective when Utilizing All Loss Terms.} Table~\ref{tab:ablation} demonstrates the impact of various terms in the loss function (Eq.~\ref{eq:optim}). Optimal performance is achieved by incorporating all loss terms during training.

\section{Additional Visualizations}
\begin{figure*}[t]
    \begin{center}
    \includegraphics[scale=0.44]{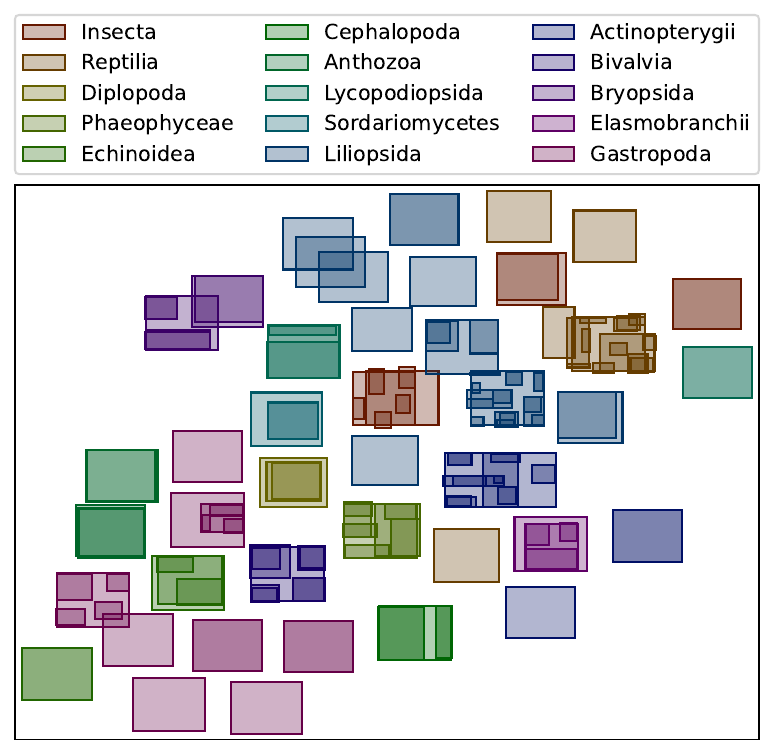}
    \includegraphics[scale=0.44]{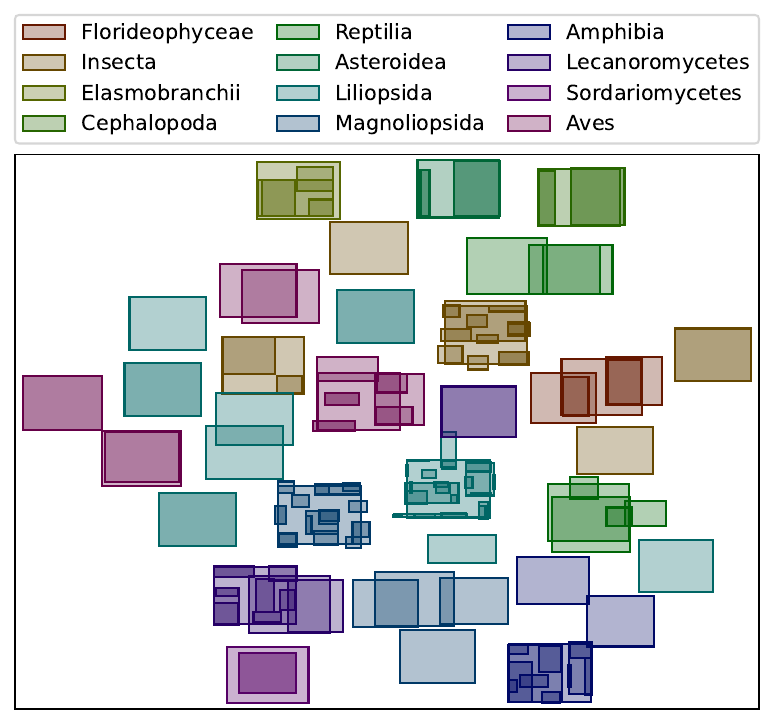}
    \includegraphics[scale=0.44]{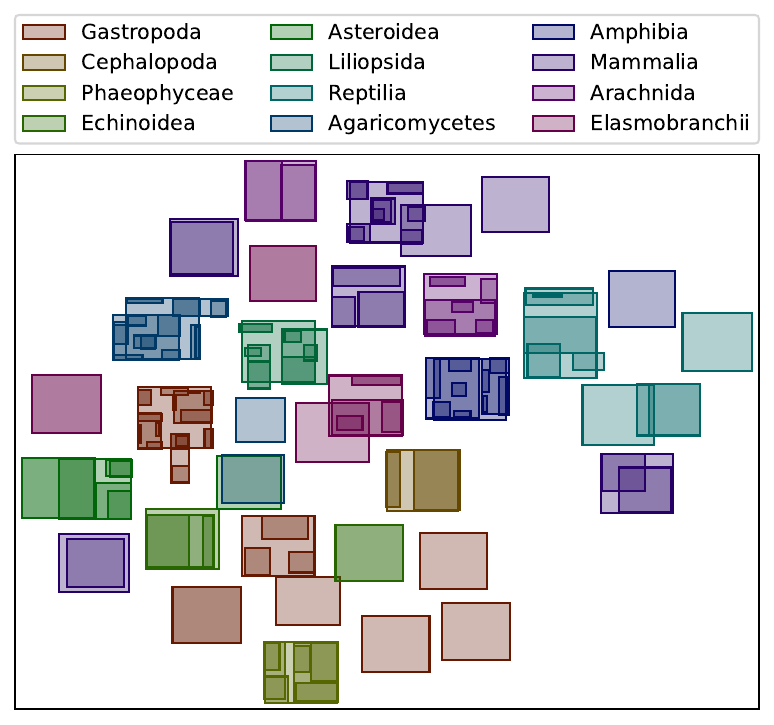}
    \end{center}
    \vspace{-0.4cm}
    \caption{\textbf{Learned embeddings for additional iNaturalist+CUB test cases}. Visualization of various test cases to evaluate \ttb. Similar to previous results, groups belonging to the same classes naturally form a hierarchy and are positioned close to each other.
    }
    \label{fig:inatcub-addtl}
\end{figure*}

\begin{figure*}[t]
    \begin{center}
    \includegraphics[scale=0.44]{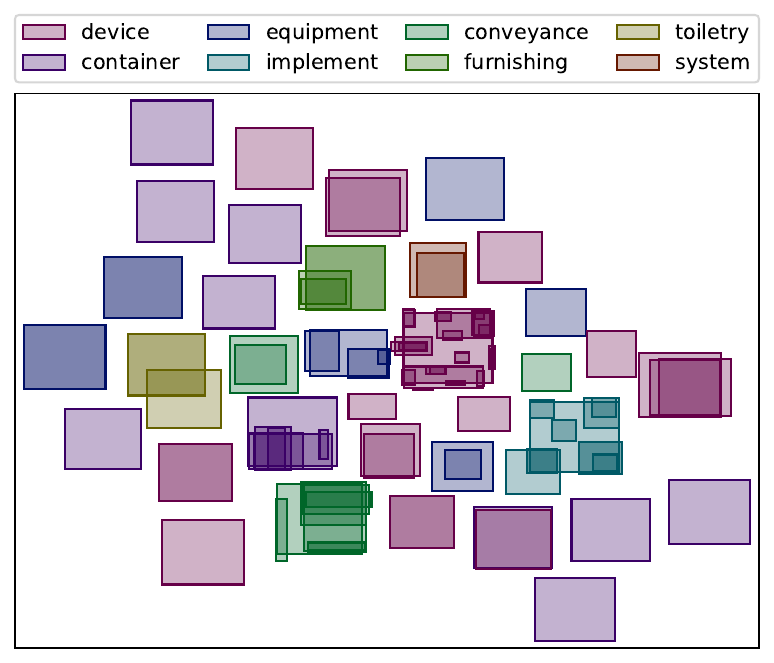}
    \includegraphics[scale=0.44]{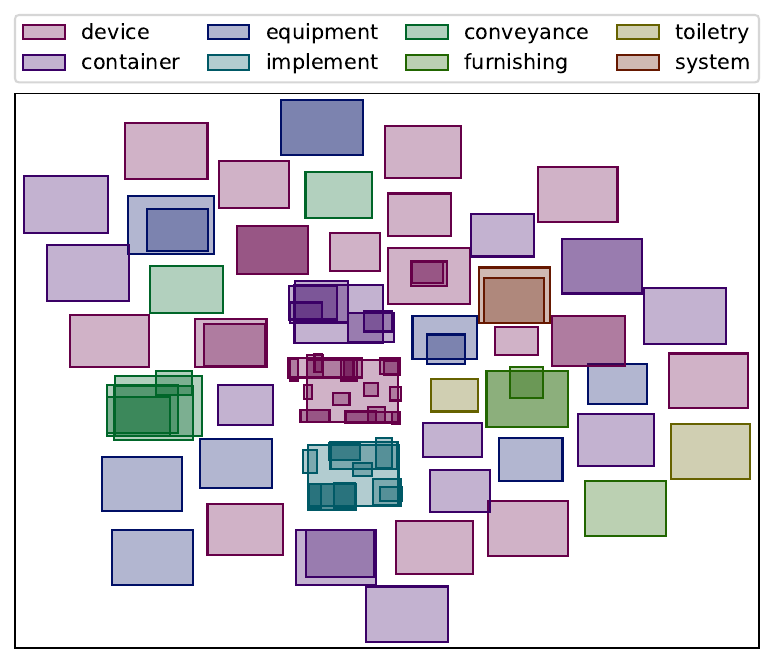}
    \includegraphics[scale=0.44]{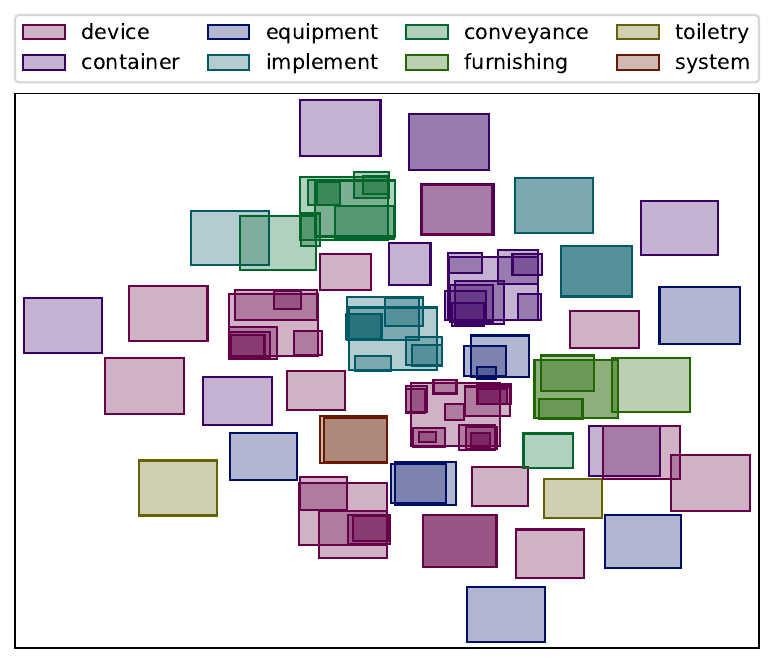}
    \end{center}
    \vspace{-0.4cm}
    \caption{\textbf{Learned embeddings for additional ImageNet test cases}. \ttb is shown to learn hierarchies in the various groups. In addition, similar objects are placed closer to each in the embedding space.
    }
    \label{fig:imagnet-addtl}
\end{figure*}

\begin{figure*}[!t]
    \begin{center}
    \includegraphics[scale=0.6]{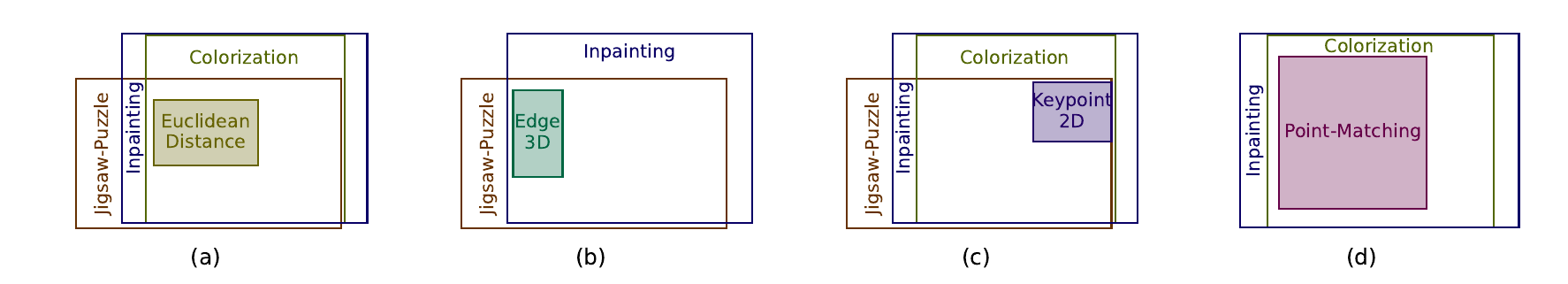}
    \end{center}
    \vspace{-6mm}
    \caption{\textbf{Additonal visualizations of tasks in Taskonomy showing source tasks that transfer well to specific target tasks (shaded)}. (a) shows that Colorization, Jigsaw Puzzle, and Inpainting are all good candidates for transfer learning onto Euclidean distance. (b), (c), and (d) show similar source task relationships for Edge-3D, Keypoints-2D, and Point Matching, respectively. It can also be observed that Inpainting transfers well to Colorization from observing subfigures (a), (c), (d).
    }
    \label{fig:taskonomy-addtl}
\end{figure*}

Fig.~\ref{fig:inatcub-addtl} and \ref{fig:imagnet-addtl} present additional visualizations for iNaturalist+CUB and ImageNet, respectively. Across various subsets of datasets, \ttb successfully learns and depicts the hierarchical relationships. Datasets that belong to the same parent category are shaded in identical colors. Fig.~\ref{fig:taskonomy-addtl} provides further visualizations for source/target task relationships within the Taskonomy benchmark, effectively illustrating source and target tasks that transfer well.

\begin{table*}
    \small
    \begin{center}
    \begin{tabular}{r | c | c | c | c  }
    \toprule
     & $\mathcal{L}_E + \mathcal{L}_D + \mathcal{L}_R$ & $-\mathcal{L}_E$ & $-\mathcal{L}_D$ & $-\mathcal{L}_R$ \\
    \hline
    iNat+CUB (F1 Score) & 84.67\% & 4.60\% & 84.56\% & 84.21\% \\
    ImageNet (F1 Score) & 90.58\% & 3.25\% & 89.02\% & 89.76\% \\
    Taskonomy ($\rho$) & 0.94 & 0.26 & 0.62 & 0.93 \\
    \bottomrule
    \end{tabular}
\caption{\textbf{Effect of Various Loss Terms on Training \ttb.} (Refer to Eq. \ref{eq:optim}). $\mathcal{L}_E$ accounts for overlap, $\mathcal{L}_D$ the distance, and $\mathcal{L}_R$ the regularization. 5D box embeddings were predicted using the base representation $\mu{CLIP}$ for iNaturalist (iNat), CUB, and ImageNet datasets; attribute-based representations were used for Taskonomy.}
    \label{tab:ablation}
    \end{center}
\end{table*}

\begin{table*}[!t]
    \small
    \begin{center}
    \begin{tabular}{l | c c c | c c c | c}
    \toprule
    \multirow{2}{*}{} & \multicolumn{3}{c|}{\textbf{ImageNet}} & \multicolumn{3}{c|}{\textbf{iNaturalist + CUB}} & {\textbf{Taskonomy}}\\
    \midrule
     & $\mu_{CLIP}$ & $[\mu,\sigma^2]_{CLIP}$ & FIM  & $\mu_{CLIP}$ & $[\mu,\sigma^2]_{CLIP}$ & FIM & Spearman's $\rho$\\
    \midrule
    \ttb (2D) & 51.94\% & 51.07\% & 27.12\%  & 53.52\% & 51.78\% & 27.21\% & 0.82 $\pm$ 0.06 \\
    \ttb (3D) & \underline{67.26\%} & \underline{66.68\%} & \underline{41.23\%}  & \underline{63.78\%} & \underline{65.38\%} & \underline{48.19\%} & \underline{0.90 $\pm$ 0.02} \\
    \ttb (5D) & \textbf{74.09\%} & \textbf{72.89\%} & \textbf{53.66\%}  & \textbf{68.11\%} & \textbf{68.60\%} & \textbf{62.61\%} & \textbf{0.92 $\pm$ 0.03} \\
    MLP & 45.44\% & 52.24\% & 31.44\% & 43.42\% & 55.08\% & 29.91\% & 0.77 $\pm$ 0.06 \\
    Linear & 12.75\% & 10.47\% & 25.90\% & 3.53\% & 5.63\% & 8.98\% & 0.70 $\pm$ 0.05 \\
    \bottomrule
    \end{tabular}
    \caption{\textbf{Performance when Trained on only 50\% of Relationships.} Even with limited data on relationships between existing datasets, \ttb shows strong performance.}
    \label{table:50perc-supervision}
    \end{center}
    \vspace{-3mm}
\end{table*}

\section{Implementation Details}
Below, we detail the configurations and hyperparameters used for extracting the base representations, training \ttb, and visualizing public datasets.

\subsection{Base Task Representations}
\subsubsection{CLIP Embeddings}
The CLIP base embeddings were generated using a ViT-H/14 network pretrained on LAION-2B. Each image was preprocessed by the provided transforms in the OpenCLIP library \cite{ilharco_gabriel_2021_5143773_openclip}. The text labels for the corresponding images were processed to be in the form ``A photo of \textsc{[CLS]}", tokenized with OpenCLIP, and encoded using the pretrained network. For iNaturalist+CUB, the labels correspond to the species of the organism in the image, whereas for ImageNet, the labels correspond to the name of the object in the image. The embeddings are taken across all images and processed as discussed in \S~\ref{sec:method}. To have a consistent setup across different base representations for iNaturalist+CUB and ImageNet, we excluded datasets that only have a single class (since \ttv only extracts embeddings for those that have multiple classes). In addition, similar to the \ttv setup, we also exclude taxa that have conflicting taxonomy between iNaturalist and CUB.

\subsubsection{\ttv Embeddings}
For each dataset, we generated embeddings using \ttv with their recommended settings, on the pretrained ResNet34 network, using the Adam optimizer over 10 epochs with $LR = 4 \times 10^{-4}$ and weight decay of $\lambda=10^{-4}.$ We used a maximum dataset size of 10,000 and 100 batches in order to generate task representations with the montecarlo method. For a given hierarchical task, we generated a dataset using a subset of classes which fell within the given hierarchy for training. For instance, if there were 20 species in a given family, our task representation for that family was the corresponding vector from \ttv trained on the 20 species within that family. We combined CUB and iNaturalist into a single dataset by merging overlapping species classes. Due to discrepancies in the taxonomy between the two datasets, we simply excluded taxa which conflicted in our experiments. Finally, since \ttv requires the training of a model, we did not generate embeddings for groupings which only contained a single element.

\subsubsection{Attribute-based Embeddings}
\label{sec:supp-taskonomy-vector}
Due to the various modalities present in Taskonomy tasks that have different types of inputs/outputs, we were unable to represent them via CLIP in a straightforward and consistent manner. As a result, we constructed a table of 15 attributes for each task which can be answered  with a \emph{yes} or \emph{no}. In order to construct unique embeddings for each task, the $i$-th attribute corresponds to the $i$-th dimension of the base representation embedding $e_i$, where $e_i=1$ if the task satisfies the attribute (i.e., when considering a specific task, the $i$-th attribute can be answered with a \emph{yes}) and $e_i=0$ otherwise. While this representation is not necessarily exhaustive, we show that \ttb can effectively generalize on unseen relationships and tasks. Future work can consider looking into available datasheets \cite{gebru2021datasheets} or dataset descriptions to construct attribute-based embeddings. Table~\ref{table:taskonomy-Qs} enumerates the attributes we considered for representing each task in Taskonomy. Table~\ref{table:taskonomy-vecs} shows the corresponding base representations for each Taskonomy task based on the enumerated attributes.

\begin{table*}[!t]
    \small
    \begin{center}
    \begin{tabular}{c | l}
    \toprule
    Dimension & Corresponding Task Attribute (answered with yes/no)\\
    \midrule
    0 & Does the task input consist of multiple images depicting different scenes?\\
    1 & Are the output spatial dimensions (height and width) the same as the input?\\
    2 & Does the task output contain geometric information?\\
    3 & Does the task output contain classification of objects?\\
    4 & Does the task require 3D knowledge?\\
    5 & Is the task a generative task?\\
    6 & Is the task output a single channel output?\\
    7 & Does the task have more than one input?\\
    8 & Does the task have more than two inputs?\\
    9 & Is the task output the result of a first order operation (e.g. edges as opposed to curvature)?\\
    10 & Does the task require generating new information?\\
    11 & Does the task require knowledge of objects?\\
    12 & Does the task require knowledge of colors/light?\\
    13 & Does the task involve camera pose estimation?\\
    14 & Does the task involve pixel alignment?\\
    \bottomrule
    \end{tabular}
    \caption{\textbf{List of attributes used to represent each task in Taskonomy and their corresponding dimension}. We did this to generate a unique representation of each task. While not necessarily exhaustive, we show \ttb can generalize using these attributes.}
    \label{table:taskonomy-Qs}
    \end{center}
    \vspace{-3mm}
\end{table*}

\begin{table*}[!t]
    \small
    \begin{center}
    \begin{tabular}{c | c | c | c | c | c | c | c | c | c | c | c | c | c | c | c}
    \toprule
    Task Name & 0 & 1 & 2 & 3 & 4 & 5 & 6 & 7 & 8 & 9 & 10 & 11 & 12 & 13 & 14\\
    \midrule
    Autoencoder & 0 & 1 & 0 & 0 & 0 & 1 & 0 & 0 & 0 & 0 & 0 & 0 & 0 & 0 & 0\\
    Colorization & 0 & 1 & 0 & 0 & 0 & 1 & 0 & 0 & 0 & 0 & 1 & 0 & 1 & 0 & 0\\
    Curvatures & 0 & 1 & 1 & 0 & 1 & 0 & 0 & 0 & 0 & 1 & 0 & 0 & 0 & 0 & 0\\
    Denoising-Autoencoder & 0 & 1 & 0 & 0 & 0 & 1 & 0 & 0 & 0 & 0 & 1 & 0 & 0 & 0 & 0\\
    Depth & 0 & 1 & 1 & 0 & 1 & 0 & 1 & 0 & 0 & 0 & 0 & 0 & 0 & 1 & 0\\
    Edge-2D & 0 & 1 & 1 & 0 & 0 & 0 & 1 & 0 & 0 & 0 & 0 & 0 & 0 & 0 & 0\\
    Edge-3D & 0 & 1 & 1 & 0 & 1 & 0 & 1 & 0 & 0 & 0 & 0 & 1 & 0 & 0 & 0\\
    Euclidean-Distance & 0 & 1 & 1 & 0 & 1 & 0 & 1 & 0 & 0 & 0 & 0 & 0 & 0 & 0 & 0\\
    Inpainting & 0 & 1 & 0 & 0 & 0 & 1 & 0 & 0 & 0 & 0 & 1 & 1 & 1 & 0 & 0\\
    Jigsaw-Puzzle & 0 & 0 & 0 & 0 & 0 & 0 & 1 & 0 & 0 & 0 & 0 & 0 & 0 & 0 & 0\\
    Keypoint-2D & 0 & 1 & 0 & 0 & 0 & 0 & 1 & 0 & 0 & 0 & 0 & 0 & 0 & 0 & 0\\
    Keypoint-3D & 0 & 1 & 0 & 0 & 1 & 0 & 1 & 0 & 0 & 0 & 0 & 0 & 0 & 0 & 0\\
    Object-Classification & 0 & 0 & 0 & 1 & 0 & 0 & 1 & 0 & 0 & 0 & 0 & 1 & 0 & 0 & 0\\
    Reshading & 0 & 1 & 1 & 0 & 1 & 0 & 1 & 0 & 0 & 0 & 0 & 0 & 1 & 0 & 0\\
    Room-Layout & 0 & 0 & 1 & 0 & 1 & 0 & 1 & 0 & 0 & 0 & 0 & 0 & 0 & 1 & 0\\
    Scene-Classification & 0 & 0 & 0 & 1 & 0 & 0 & 1 & 0 & 0 & 0 & 0 & 0 & 0 & 0 & 0\\
    Segmentation-2D & 0 & 1 & 0 & 0 & 0 & 0 & 0 & 0 & 0 & 0 & 0 & 0 & 0 & 0 & 0\\
    Segmentation-3D & 0 & 1 & 0 & 0 & 1 & 0 & 0 & 1 & 1 & 0 & 0 & 0 & 0 & 0 & 0\\
    Segmentation-Semantic & 0 & 1 & 0 & 1 & 0 & 0 & 0 & 0 & 0 & 1 & 0 & 1 & 0 & 0 & 0\\
    Surface-Normal & 0 & 1 & 1 & 0 & 1 & 0 & 0 & 0 & 0 & 0 & 0 & 0 & 0 & 0 & 0\\
    Vanishing-Point & 0 & 0 & 1 & 0 & 1 & 0 & 1 & 0 & 0 & 0 & 0 & 0 & 0 & 0 & 0\\
    Pairwise-Nonfixated-Camera-Pose & 1 & 0 & 1 & 0 & 1 & 0 & 1 & 1 & 0 & 0 & 0 & 0 & 0 & 1 & 0\\
    Pairwise-Fixated-Camera-Pose & 1 & 0 & 1 & 0 & 1 & 0 & 1 & 1 & 0 & 0 & 0 & 0 & 0 & 1 & 1\\
    Triplet-Fixated-Camera-Pose & 1 & 0 & 1 & 0 & 1 & 0 & 1 & 1 & 1 & 0 & 0 & 0 & 0 & 1 & 1\\
    Point-Matching & 1 & 0 & 0 & 0 & 0 & 0 & 1 & 1 & 0 & 0 & 0 & 0 & 0 & 0 & 1\\
    \bottomrule
    \end{tabular}
    \caption{\textbf{Base representation used for each task for the attribute-based embedding}. The $i$-th dimension signifies a yes/no response to the question in Table \ref{table:taskonomy-Qs}, where $e_i=1$ for yes, and $e_i=0$ otherwise.}
    \label{table:taskonomy-vecs}
    \end{center}
    \vspace{-3mm}
\end{table*}

\subsection{\ttb Training Details}

The box embedding library from \cite{chheda2021box} was used to create instances of boxes. To compute the loss in Eq.~\ref{eq:box_dist}, we use a volume temperature of $0.1$, and intersection temperature of $0.0001$. The model for \ttb uses a 3-layer MLP with two linear layer heads, and is trained using the loss in Eq.~\ref{eq:optim}. The Adam optimizer is used with $LR = 1 \times 10^{-3}$.

For hierarchical datasets, we train all models (including baselines) on at least 150 datasets at a time, with 18,000 relationships used for training, 2,250 for validation, and the remaining 2,250 relationships for evaluation. An unseen collection of 100 datasets (15,000 relationships with existing datasets) are used as novel dataset evaluations. The performance is averaged over multiple instances of training and testing from randomly sampled datasets and relationships.

For the Taskonomy benchmark, models are also trained on a subset of randomly sampled relationships $\mathcal{R}$ within existing datasets $\mathcal{D}_E$. Eq.~\ref{eq:optim} is optimized such that relationships between low dimension embeddings $z$ match the task affinities provided by Taskonomy \cite{zamir2018taskonomy}. Since Taskonomy is limited to 25 vision tasks, only 3 could be used for evaluating the performance on novel datasets. The rest of the vision tasks were used for training and validation. Similar to the hierarchical datasets, multiple instances were sampled for performance evaluation. At the same time, while task affinities from \cite{zamir2018taskonomy} were already normalized in the range $[0,1]$, the distribution is skewed to the right. The task affinities are re-scaled using Eq.~\ref{eq:tanh-taskonomy} to normalize the distribution where $x$ is the task affinity value, $x'$ is the re-scaled task affinity value, and $k$ is a hyperparameter we set to 50. We train all models using the re-scaled task affinities, and convert it back to the original scale for evaluation.

\begin{equation}
    x' = \frac{\exp{(kx)} - \exp{(-kx)}}{\exp{(kx)} + \exp{(-kx)}}
    \label{eq:tanh-taskonomy}
\end{equation}

\subsection{Public Dataset Visualization}
Since no ground truth relation labels are available for training the model on public datasets, we use the following ``soft'' labels for defining relationships between pairs of datasets. The soft labels are defined as the asymmetric overlap (similar to Eq.~\ref{eq:box_dist}) between the base representation of datasets. For the Hugging Face visualization, we use CLIP embeddings as the base representation: for each dataset, we sample $N$ image-label pairs where $N \leq 10,000$, then we embed the images and the labels using CLIP to produce $N$ embeddings per dataset. To get the soft overlap value between two datasets $so(\mathcal{D}_i,\mathcal{D}_j)$, we use Eq. \ref{eq:softoverlap} where $\mathcal{E}_i=\{e_{i}^{(1)}, e_{i}^{(2)}, \ldots\, e_{i}^{(N)}\}$ is the set of image-label embeddings for dataset $\mathcal{D}_i$, $co(\mathcal{D}_i,\mathcal{D}_j)$ is the count of overlapping embeddings of $\mathcal{D}_i$ with respect to $\mathcal{D}_j$, and $|\mathcal{E}_i|$ is the number of embeddings in $\mathcal{E}_i$.

\begin{equation}
    so(\mathcal{D}_i,\mathcal{D}_j) = \frac{co(\mathcal{D}_i,\mathcal{D}_j)}{|\mathcal{E}_i|}
    \label{eq:softoverlap}
\end{equation}

In Eq.~\ref{eq:softoverlap}, $co(\mathcal{D}_i,\mathcal{D}_j)$ counts the number of embeddings $e_{i}^{(k)} \in \mathcal{E}_i$ that satisfy $d_{ij}^{(k)} < t_j$, where $d_{ij}^{(k)}$ is the minimum euclidean distance of $e_{i}^{(k)}$ among all embeddings $e_{j}^{(k)} \in \mathcal{E}_j$ of dataset $\mathcal{D}_j$, and $t_j$ is the average euclidean distance $d_{euc}(\cdot, \cdot)$ between any two embeddings in $\mathcal{D}_j$. Eq.~\ref{eq:count-overlap} shows how $co(\mathcal{D}_i,\mathcal{D}_j)$ is computed where $[\cdot]$ is an indicator function that evaluates to $1$ if the expression inside the brackets is true, and $0$ otherwise. $N=|\mathcal{E}_i|$ which is the number of image-label embeddings in dataset $\mathcal{D}_i$.

\begin{equation}
    co(\mathcal{D}_i,\mathcal{D}_j) = \sum_{k=1}^N \ [d_{ij}^{(k)} < t_j]
    \label{eq:count-overlap}
\end{equation}

Eq.~\ref{eq:dik} and Eq.~\ref{eq:thresh-softoverlap} show how $d_{ij}^{(k)}$ and $t_j$ are computed. $M=|\mathcal{E}_j|$ which is the number of image-label embeddings in dataset $\mathcal{D}_j$. 
\begin{equation}
    d_{ij}^{(k)} = \min_{e_{j}^{(l)} \in \mathcal{E}_j}  d_{euc}\left(e_{i}^{(k)}, e_{j}^{(l)} \right) 
    \label{eq:dik}
\end{equation}
\begin{equation}
    t_{j} = \frac{2}{M(M-1)} \sum_{u=1}^M \sum_{v=u+1}^M d_{euc}\left(e_{j}^{(u)}, e_{j}^{(v)}\right)
    \label{eq:thresh-softoverlap}
\end{equation}

The soft overlaps between all pairs of datasets are computed and used as supervision to train \ttb. Note that similar to Eq. \ref{eq:box_dist}, Eq. \ref{eq:softoverlap} is also an asymmetric measure of similarity between two datasets. The input to \ttb uses the average CLIP embedding per dataset (Eq.~\ref{eq:clip_mu}), and the model is trained using Eq. \ref{eq:optim}. We additionally include a loss term $\mathcal{L}_A$ that encourages the area of the box embedding $z_i$ to correspond to the size of the dataset $|\mathcal{D}_i|$ (i.e., the number of samples available in the dataset). Eq. \ref{eq:loss-area} shows how the loss is computed. $\mathcal{L}_A$ is added to the objective function discussed in Eq. \ref{eq:optim}.

\begin{equation}
    \mathcal{L}_A = \text{MSE} \left( \text{vol}(z_i), |\mathcal{D}_i| \right)
    \label{eq:loss-area}
\end{equation}

Results on image classification datasets from Hugging Face are shown in Fig. \ref{fig:huggingface}. The same method can also be applied to other datasets where only images and corresponding labels are available.

\section{Descriptions of Taskonomy Tasks}
\label{sec:taskonomy-task-desc}
Below, we provide a brief description of each task. For a more complete definition, we refer to Taskonomy~\cite{zamir2018taskonomy}. 
\begin{enumerate}
    \item Autoencoder: Using an encoder-decoder neural network that takes an input image, distills it to a single vector representation, then reconstructs the image.
    \item Colorization: Selecting pixel color assignments for a black and white image.
    \item Curvatures: Given an image, identify the degree of curvature of the physical object on each pixel.
    \item Denoising-Autoencoder: Denoise an image using an encoder-decoder structure.
    \item Depth: Find the z-buffer depth of objects in every pixel of an image.
    \item Edge-2D: Identify strong boundaries in the image.
    \item Edge-3D: Find occlusion edges, where an object in the foreground obscures things behind it.
    \item Euclidean-Distance: For each pixel, find the distance of the object to the camera.
    \item Inpainting: Given a part of an image, reconstruct the rest.
    \item Jigsaw-Puzzle: Given different parts of an image, reassemble the parts in order.
    \item Keypoint-2D: Find good pixels in the image which are distinctive for feature descriptors.
    \item Keypoint-3D: Find good pixels like in Keypoint-2D, but using 3D data and ignoring distracting features such as textures. 
    \item Object-Classification: Assign each image to an object category.
    \item Reshading: Given an image, generate a reshaded image which results from a single point light at the origin.
    \item Room-Layout: Given an image, estimate the 3D layout of the room.
    \item Scene-Classification: Assign a scene category to each image.
    \item Segmentation-2D: Group pixels within an image, based on similar-looking areas.
    \item Segmentation-3D: Group pixels within an image, based on both the image and depth image and surface normals.
    \item Segmentation-Semantic: Assign each pixel to an object category.
    \item Surface-Normal: Assign each pixel a vector representing the surface normal. 
    \item Vanishing-Point: Identify the vanishing points within an image.
    \item Pairwise-Nonfixated-Camera-Pose: Identify the 6 degrees of freedom relative camera pose between two images.
    \item Pairwise-Fixated-Camera-Pose: Identify the 5 degrees of freedom relative camera pose between two images which share a pixel center.
    \item Triplet-Fixated-Camera-Pose: Identify the relative camera poses between three images.
    \item Point-Matching: Given two images and one point, identify the matching point in the other image.
\end{enumerate}



\end{document}